\pdfoutput=1

\documentclass[runningheads]{llncs}
\usepackage[colorlinks,allcolors=green]{hyperref}

\usepackage{latexsym}
\usepackage{lipsum}
\usepackage[T1]{fontenc}

\usepackage[utf8]{inputenc}

\usepackage{enumerate}
\usepackage{graphicx}
\usepackage{booktabs}
\usepackage[skip=0pt]{caption}
\usepackage{subcaption}
\usepackage{graphics, amsmath}
\usepackage{amssymb}
\usepackage{multirow}

\usepackage{tabularx}
\usepackage{pifont}
\usepackage{colortbl}
\usepackage{xspace}
\usepackage{xcolor}
\usepackage[most]{tcolorbox}
\usepackage{makecell}
\usepackage{adjustbox}
\usepackage[inline]{enumitem}

\usepackage{arydshln}
\usepackage{acronym}

\usepackage{wrapfig}

\usepackage{floatrow}
\newfloatcommand{capbtabbox}{table}[][\FBwidth]

\usepackage{blindtext}

\newcommand\blfootnote[1]{%
  \begingroup
  \renewcommand\thefootnote{}\footnote{#1}%
  \addtocounter{footnote}{-1}%
  \endgroup
}

\newtcolorbox[list inside=prompt,auto counter,number within=section]{prompt}[1][]{
    colbacktitle=black!60,
    coltitle=white,
    fontupper=\footnotesize,
    boxsep=5pt,
    left=0pt,
    right=0pt,
    top=0pt,
    bottom=0pt,
    boxrule=1pt,
    #1,
}

\newcommand{\change}[1]{#1}

\newcommand{\dataname}{KI-QFS}

\newcommand{\querysum}{\textsc{QuerySum}}
\newcommand{\margesum}{\textsc{MargeSum}}

\newcommand{\corpus}{\textsc{Corpus}}
\newcommand{\origin}{\textsc{Origin}}
\newcommand{\corpusint}{\corpus{}\textsubscript{Int}}
\newcommand{\corpusext}{\corpus{}\textsubscript{Ext}}
\newcommand{\corpusaug}{\corpus{}\textsubscript{Aug}}

\begin{document}

\title{Beyond Relevant Documents: A Knowledge-Intensive Approach for Query-Focused Summarization using Large Language Models}
\titlerunning{Knowledge-Intensive Query-Focused Summarization}

\author{
Weijia Zhang\inst{1}$^*$ \and Jia-Hong Huang\inst{1}$^*$ \and Svitlana Vakulenko\inst{2} \and Yumo Xu\inst{3} \and Thilina Rajapakse\inst{1}  \and Evangelos Kanoulas\inst{1}}

\authorrunning{W. Zhang et al.}
\institute{University of Amsterdam\inst{1}, Amazon Alexa AI\inst{2}, University of Edinburgh\inst{3}
}

\maketitle

\begin{abstract}

Query-focused summarization (QFS) is a fundamental task in natural language processing with broad applications, including search engines and report generation. However, traditional approaches assume the availability of relevant documents, which may not always hold in practical scenarios, especially in highly specialized topics. To address this limitation, we propose a novel knowledge-intensive approach that reframes QFS as a knowledge-intensive task setup. 
This approach comprises two main components: a retrieval module and a summarization controller. The retrieval module efficiently retrieves potentially relevant documents from a large-scale knowledge corpus based on the given textual query, eliminating the dependence on pre-existing document sets. 
The summarization controller seamlessly integrates a powerful large language model (LLM)-based summarizer with a carefully tailored prompt, ensuring the generated summary is comprehensive and relevant to the query.
To assess the effectiveness of our approach, we create a new dataset, along with human-annotated relevance labels, to facilitate comprehensive evaluation covering both retrieval and summarization performance. 
Extensive experiments demonstrate the superior performance of our approach, particularly its ability to generate accurate summaries without relying on the availability of relevant documents initially. This underscores our method's versatility and practical applicability across diverse query scenarios.

\end{abstract}

\blfootnote{$^*$Equal contribution.}

\begin{figure}[t]
\centering
\includegraphics[width=0.985\textwidth]{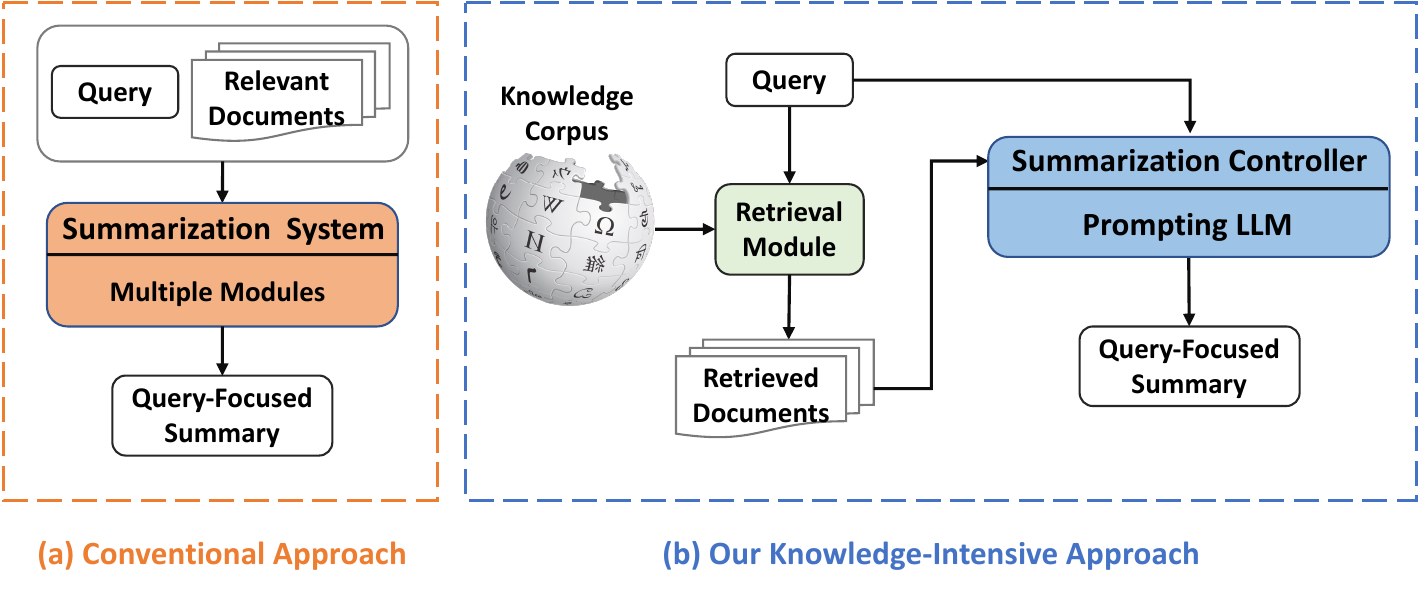}
\caption{The Comparison between (a) the conventional approach and (b) our knowledge-intensive approach. The conventional one assumes relevant documents are available. We aim to retrieve such documents from a large knowledge corpus.}
\label{fig:comparison}
\end{figure}

\section{Introduction}

Query-focused summarization (QFS) is a pivotal task with wide-ranging applications, spanning fields such as search engines and report generation~\cite{ma2023mdssurvey}. It involves analyzing a textual query alongside a collection of relevant documents to automatically produce a textual summary closely aligned with the query~\cite{dang-2006-duc,xu-lapata-2020-coarse,xu-lapata-2021-generating,liu2024querysum}. This process aims to offer users relevant insights in a condensed format by applying information compression techniques to the provided document set~\cite{li2017salience,laskar-etal-2020-wsl,xu-lapata-2021-generating}. 
The conventional approach to this task assumes the availability of a set of relevant documents and creates the summary based on the information within this document set. However, in practical scenarios, this assumption may not always hold.

Consider highly specialized or niche topics as an illustration. In fields such as advanced scientific research, specialized technology, or obscure hobbies, documentation may be limited or fragmented, particularly for cutting-edge or esoteric subjects.
Similarly, think about future predictions or speculations. Queries related to future events, trends, or predictions may lack existing documents as they entail speculation or anticipation of events yet to unfold. For instance, questions about the outcome of upcoming elections, the impact of emerging technologies, or the trajectory of financial markets might not be documented until after these events have occurred. In the aforementioned practical scenarios, the traditional setup of QFS with compression techniques may not work well due to the scarcity or absence of relevant documents.

To tackle this challenge, our approach reframes QFS as a knowledge-intensive (KI) task setup~\cite{petroni-etal-2021-kilt}. Unlike the traditional setup where relevant documents are assumed to be readily available, our knowledge-intensive approach only assumes access to a large-scale knowledge corpus.
Our proposed approach comprises a retrieval module and a summarization controller.
Given a textual query as the initial input, the retrieval module returns potentially relevant documents from the knowledge corpus, eliminating the dependence on pre-existing document sets. These retrieved documents, along with the query, are then forwarded to the summarization controller. 
The summarization controller harmoniously merges an advanced large language model (LLM)-powered summarizer with a carefully designed prompt, guaranteeing the produced summary is both comprehensive and directly addresses the query.
To illustrate the aforementioned process, we make a comparison between the conventional approach and our proposed approach in Figure~\ref{fig:comparison}.

To validate the effectiveness of our proposed approach within the KI setup, we have introduced a specialized dataset by combining existing QFS datasets~\cite{dang-2006-duc} with three various knowledge corpora. We also enhance the dataset with human-annotated relevance labels to facilitate comprehensive evaluation for both retrieval and summarization performance. This dataset serves as the foundation for a fair evaluation of our approach and baseline models. 
Extensive experiments conducted in this work confirm the effectiveness of our proposed method, surpassing baseline models and showcasing its superiority in generating query-focused summaries. Notably, our approach stands out for its ability to generate accurate summaries without relying on the availability of relevant documents at the initial input stage. This underscores the versatility of our method in addressing diverse query scenarios, further validating its practical applicability.

\vspace{+3pt}
\noindent The primary contributions of this work can be summarized as follows:
\begin{enumerate}[label=\arabic*),nosep]
    \item \textbf{Introduction of a Knowledge-Intensive Approach}: We introduce a new knowledge-intensive approach for QFS, departing from the conventional paradigm that relies on the availability of relevant documents in the initial input stage. This innovative method involves the direct retrieval of pertinent documents from an extensive knowledge corpus, leveraging the provided textual query as the basis.

    \item \textbf{Development of a Summarization Controller}: We propose a summarization controller that orchestrates the process of generating query-focused textual summaries using an integrated LLM-based summarizer with a specially designed prompt. This controller streamlines the summarization process, ensuring alignment with the query intent.
    
    \item \textbf{Creation of a Specialized Dataset}: To assess the effectiveness of the proposed approach, we introduce a new dataset explicitly designed for knowledge-intensive QFS. This dataset, along with human-annotated relevance labels, serves as a standardized benchmark for evaluating both retrieval and summarization performance.

     \item \textbf{Demonstration of Superior Performance}: We present extensive experimental results showcasing the superior performance of the proposed approach compared to baseline models. Particularly noteworthy is the approach's capability to generate accurate summaries without the prerequisite of available relevant documents at the initial input stage. This feature underscores the method's versatility and practical utility across a range of query scenarios.
    
\end{enumerate}

\section{Related Work}

In this section, we cover three lines of research: query-focused summarization, knowledge-intensive language tasks, and open-domain summarization, all related to our study.

\subsection{Query-Focused Summarization}

Query-focused summarization aims to generate a summary tailored to a given query from a set of topic-related contexts. For instance, the contexts can be texts~\cite{dang-2006-duc,liu2024querysum}, tables~\cite{zhao-etal-2023-qtsumm,zhang2024qfmts}, or videos~\cite{huang2020query,huang2021gpt2mvs,huang2021deepopht,huang2023causalainer,huang2023improving,huang2023query}. We mainly focus on textual contexts.
\change{Conventional QFS methods primarily focused on sentence-level extraction techniques to eliminate redundant information~\cite{li2017salience,laskar-etal-2020-wsl,xu-lapata-2020-coarse,xu-lapata-2021-generating}. However, they heavily rely on pre-existing relevant documents and are typically developed for a limited set of closed-domain documents ($\sim$50 documents). In contrast to all previous work, our work does not assume the existence of relevant documents to the given query. Instead, our approach handles open-domain document-level retrieval, aiming to identify a small set of relevant documents from a pool of millions of candidates, accommodating more diverse topics and contexts.} 

Another line of research aims to automatically create larger QFS datasets to alleviate data scarcity, utilizing information retrieval or clustering methods~\cite{nema-etal-2017-diversity,kulkarni2020aquamuse,pasunuru2021qfs,liu2024querysum}.%
However, these datasets lack a standardized benchmark to assess the retrieval performance, as they lack relevance annotations for retrieved documents. In contrast, our dataset includes human annotations for the relevance of retrieved documents. This facilitates retrieval evaluation and provides insights into understanding the effects of retrieval errors on summarization performance.

\subsection{Knowledge-Intensive Tasks}

Knowledge-intensive language tasks~\cite{petroni-etal-2021-kilt} involve addressing user needs by leveraging a large-scale knowledge corpus. 
These tasks produce various types of outputs. For instance, they can generate short-form factual answers \cite{chen-etal-2017-reading,huang2019novel,huang2024novel,huang2024optimizing,zhu2024enhancing}, dialog responses~\cite{dinan2019wow}, or face-checking predictions~\cite{thorne-etal-2018-fever}.
One of the most related KI tasks is a long-form question answering (LFQA)~\cite{fan-etal-2019-eli5,stelmakh-etal-2022-asqa,zhang2024towards}, where the goal is to provide comprehensive answers to given questions. 
\change{Compared to QFS, the key difference between LFQA and QFS lies in their objectives and the role of input documents. LFQA provides detailed answers to complex questions, whereas documents are optional external sources to enhance answer quality~\cite{fan-etal-2019-eli5,stelmakh-etal-2022-asqa}. For instance, recent studies~\cite{amplayo-etal-2023-query} investigated the effectiveness of LLMs in LFQA without relying on external documents. Additionally, generating long-form answers does not necessarily require multiple documents. In contrast, QFS requires generating a concise summary from multiple documents, tailored to a specific query. In this case, the relevant documents are essential inputs to generate the summary~\cite{xu-lapata-2020-coarse,laskar-etal-2020-wsl}.}

Furthermore, existing LFQA tasks either struggle with accurately grounding their answers in the provided documents~\cite{krishna-etal-2021-hurdles} or they heavily rely on the inherent ambiguity present in questions~\cite{min-etal-2020-ambigqa}. In comparison, our work has two significant advantages: 
\begin{enumerate*}[label=\arabic*)]
    \item Our approach ensures generated query-focused summaries are firmly based on the content of the retrieved documents; and
    \item The queries are unambiguous and accurately reflect complex information needs in QFS, resulting in naturally longer summaries to address various information needs adequately.
\end{enumerate*}

\subsection{Previous Attempts on Open-Domain Summarization}

Our preliminary research, first presented in the arXiv and non-archival workshop versions~\cite{zhang2021scaling,zhang2023tackling}, marks the initial phase of investigating the complexities of open-domain QFS. These findings have yet to be officially published in any conference proceedings or archival workshops. 
Building upon this work, recent studies~\cite{giorgi-etal-2023-open,zhou2023odsum} have embarked on similar investigations. They either leverage multi-document summarization (MDS) datasets~\cite{fabbri-etal-2019-multi} or adapt query-based single-document summarization datasets~\cite{zhong-etal-2021-qmsum,wang-etal-2022-squality}. In comparison, this work has two significant differences: 
\begin{enumerate*}[label=\arabic*)]
    \item Previous studies lack human annotation for the relevance of the retrieved documents, raising concerns about the effectiveness of their evaluation of retrieval models. Instead, we conduct human annotations to ensure high-quality relevance annotations, thus enabling more accurate evaluation of retrieval models; and
    \item Previous studies primarily focused on internal documents within specific domains, such as news or medicine, limiting their analysis to closed domains. In contrast, our study takes a broader approach by considering both internal documents and a significantly larger external knowledge corpus, e.g., Wikipedia. This broader scope enhances the applicability of our findings to real-world scenarios.
\end{enumerate*}

\section{Knowledge-Intensive Approach}

In this section, we formulate our knowledge-intensive task setup and detail our proposed approach.

\subsection{Task Formulation}
\label{subsec:task}

To better understand the difference between traditional QFS and our knowledge-intensive setup, we first describe the traditional QFS and then formulate our proposed knowledge-intensive setup.
\subsubsection{Traditional QFS.}

Let the tuple $(q, \mathcal{D}, s)$ denotes an example in the traditional QFS task, where $q$ is a given query, $\mathcal{D}=\{d_1, \dotsc, d_m\}$ is a set of relevant documents and $s$ is the gold summary for the document set $\mathcal{D}$ and the query $q$. The goal of QFS is to produce the query-focused summary given the documents and the query as input: $(q, \mathcal{D}) \rightarrow s$. 
It is worth noting that $\mathcal{D}$ is collected manually by human annotators and usually includes tens of documents. 

\subsubsection{Knowledge-Intensive QFS.}

In this work, we reframe the QFS task to a knowledge-intensive task setup as $q \rightarrow s$. As shown in Figure \ref{fig:comparison}, we do not rely on the relevant document set $\mathcal{D}$. Instead, we assume the access to a large-scale knowledge corpus: $\mathcal{K}=\{d_1, \dotsc\, d_n\}$. In practice, $\mathcal{K}$ consists of millions of documents, i.e., $n \gg m$.
It is worth noting that our goal remains to generate the query-focused summary $s$ with respect to the query $q$. However, different from the traditional setup, there is no guarantee that the provided documents are relevant to the query. Instead, there are millions of irrelevant documents in the knowledge corpus. Thus, an effective information retrieval model is necessary to supplement the extended setup, which is described below.

\subsection{Methodology}
\label{subsec:method}

Our approach includes two components: a retrieval module and a summarization controller, as illustrated in Figure~\ref{fig:comparison}. The retrieval module returns top-$k$ documents from the knowledge corpus based on the given query. Then the summarization controller takes the query and retrieved documents as the inputs to generate a query-focused summary.

\subsubsection{Retrieval Module.}

Given a query $q$ and a document $d_i$ from the knowledge corpus $\mathcal{K}$, the retrieval module first estimates the relevance $Rel(q, d_i)$ between $q$ and $d_i$ and then ranks all documents based on their relevance scores. Typically, we only consider top-$k$ retrieved documents. There are two main categories of retrieval models: sparse and dense models~\cite{zhao2024dense}. The former estimates relevance between the query and the document using weighted counts of their overlapping terms. The latter first transforms both the query and the document into vectors within an embedding space. The relevance is then computed using the dot product of their respective vectors. In this work, we explore representative sparse and dense models, which are BM25~\cite{robertson2009bm25} and Dense Passage Retrieval (DPR)~\cite{karpukhin-etal-2020-dense}, respectively. 

\subsubsection{Summarization Controller.}

After we obtain top-$k$ retrieved documents from the retrieval module, we feed the query and the documents into the summarization controller to generate a query-focused summary. As large language models (LLMs), such as GPT-3.5~\cite{ouyang2022rlhf}, have exhibited remarkable generation ability in many text generation tasks~\cite{laskar-etal-2023-systematic}. \change{Inspired by Chain-of-thought prompting methods~\cite{wei2022cot,kojima2022zerocot}, we introduce a multi-step summarization controller. As shown in Prompt \ref{prompt:controller}, the controller involves two primary steps: identifying query-relevant information and generating a controllable summary. First, to ensure the summary's relevance to the query, we explicitly instruct the LLM to identify query-relevant information within the documents. It is worth noting that the LLM can determine information at various levels, including phrases, sentences, and paragraphs. Subsequently, we instruct the LLM to write a controllable summary that meets the length limit of 250 words based on the query-relevant information. This ensures the summary is sufficiently comprehensive and relevant to the query. Moreover, we include few-shot demonstrations in the prompt to enable in-context learning for the LLM, which has been proven to be effective in many natural language processing tasks~\cite{brown2020gpt3}.}

\begin{prompt}[title={Prompt \thetcbcounter: Summarization Controller}, label=prompt:controller]
\textbf{Instruction:} You will be given a query and a set of documents. Your task is to generate an informative, fluent, and accurate query-focused summary. \change{To do so, you should obtain a query-focused summary step by step.} \\

\change{\textbf{Step 1: Query-Relevant Information Identification}} \\
\change{In this step, you will be given a query and a set of documents. Your task is to find and identify query-relevant information from each document. This relevant information can be at any level, such as phrases, sentences, or paragraphs.} \\

\change{\textbf{Step 2: Controllable Summarization}} \\
\change{In this step, you should take the query and query-relevant information obtained from Step 1 as inputs. Your task is to summarize this information. The summary should be concise, include only non-redundant, query-relevant evidence, and be approximately 250 words long.} \\

\textbf{Demonstrations:} \\
Few-shot human-written demonstrations. \\
\\
\textbf{Query:} \{\textit{Input query}\} \\
\textbf{Documents:} \{\textit{Retrieved documents}\}
\end{prompt}

\section{\dataname{} Dataset}

In this section, we describe dataset collection and relevance annotation process for our knowledge-intensive setup. The dataset includes a collection of query-summary pairs and three alternatives of knowledge corpora. The relevance annotation aims to facilitate retrieval evaluation.

\subsection{Dataset Collection}

\subsubsection{Collecting Query-Summary Pairs.}

We build our dataset on the top of query-summary pairs $(q, s)$ on existing QFS resources. 
Specifically, we adopt DUC 2005-07~\cite{dang-2006-duc}, three standard QFS benchmark datasets.\footnote{We take DUC as an example, but nothing prohibits exploring other QFS resources.}
The DUC datasets consist of three subsets collected for the Document Understanding Conferences (DUC) from 2005 to 2007. Each subset contains 45-50 clusters. Each cluster (or a topic) contains a query, a set of topic-related documents, and multiple reference summaries. As relevant documents are inaccessible in our dataset, we only collect query-summary pairs first. The relevant documents will be used to build a knowledge corpus, which is described below.
For data split, we follow previous work~\cite{li2019document,laskar-etal-2020-wsl} to use the pairs of the first two years (2005-2006) as the training set and randomly select 10\% from the training set as the validation set. We leave the third subset (2007) as the test set. Finally, the training set, validation set, and test set contain 90, 10, and 45 examples, respectively.

\subsubsection{Building Knowledge Corpora.}

We explore three alternatives of knowledge corpora \corpus{} for the query-summary pairs above. Specifically, we first follow the standard data processing for large-scale knowledge sources \cite{karpukhin-etal-2020-dense} to collect documents, where we split each origin document into non-overlapping context documents with 100 words maximum.
\footnote{Some studies~\cite{karpukhin-etal-2020-dense} also use passages to name the processed text chunks, but we adhere to the term ``documents'' to maintain the coherence of the paper.}
The knowledge corpora are as follows:
\begin{enumerate*}[label=\arabic*)]
    \item The first knowledge corpus, \corpusint{}, is an internal collection, where we take all clusters of DUC documents from the three years to form this collection, which results in about 32K documents in total. As the queries have an explicit connection with the documents, this collection can be considered an in-domain knowledge corpus, where relevant documents are relatively easy to find.
    \item However, as our main goal is to explore knowledge-intensive QFS on the large-scale knowledge corpus, we consider the second external knowledge corpus named \corpusext{}. We use the Wikipedia dump in the KILT benchmark~\cite{petroni-etal-2021-kilt} to form this corpus, resulting in about 21 million documents in total. 
    \item However, as \corpusext{} is a Wiki-based corpus, there is no guarantee that the collection contains sufficient evidence to answer the query since the reference summary is derived from the content of the original DUC documents. To this end, we build an augmented knowledge corpus called \corpusaug{} by combining the previous two collections. We merge all the documents of \corpusint{} and \corpusext{} into this collection ensuring that summaries can be grounded in the collection and that the collection is large-scale to make the task challenging.
\end{enumerate*}

\subsection{Relevance Annotation}
\label{subsec:retrilabel}

As the retrieval process is necessary for our setup, it is important to evaluate the performance of retrieval models so we can better analyze their effect on summarization.
However, the DUC datasets do not come with relevance labels to designate which document provides evidence for the final summary. A possible solution to that would be to do a lexical match between the individual summary sentences and the available document collection, however, such an automatic evaluation has its disadvantages (e.g. a summary sentence may appear in a document but outside the right context, lexical overlap may pay attention to insignificant words in the summary, and there may be a lexical gap between semantically similar sentences). 
Therefore, we collect human annotations through Amazon Mechanical Turk (MTurk) to create our evaluation set. 

As the main goal of the annotation process is to obtain relevant documents that support the summary, we begin with filtering out irrelevant documents, which are the majority of millions of candidate documents in the knowledge corpus.
Following the TREC document retrieval task \cite{craswellC2020trecdeep}, we apply retrieval models to choose top-ranked documents, as discussed in subsection \ref{subsec:method}. We first perform both BM25 and DPR on \corpusint{} and \corpusext{}, and construct top-50 pools, which result in a maximum of 200 candidate documents for each query. As there are 45 queries in our test set, we end up with 7761 query-document pairs to be annotated. We then design a web annotation interface shown to MTurk workers, which contains a query, a reference summary, and at most five documents. We ask workers to label whether a document contains either part of the summary or evidence to a query, adapting the annotation instructions in \cite{craswellC2020trecdeep}. 
The judgments are on a 4-point scale from 0 to 3:
\begin{enumerate*}[label=(\roman*)]
    \item \textbf{0-irrelevant}: the document has nothing to do with the query;
    \item \textbf{1-weakly related}: the document seems related to the query but fails to contain any evidence in the summary; 
    \item \textbf{2-related}: the document provides unclear information related to the query, but human inference may be needed; and
    \item \textbf{3-relevant}: the document explicitly contains evidence that is part of the summary.
\end{enumerate*}
Prior to the collection of the annotations we performed two pilot studies to improve our annotation interface and the instructions. We collect three annotations per document and use a majority vote to determine the final relevance label. The distribution of relevance labels is 837 relevant, 5811 related, 1097 weakly related, and 16 irrelevant documents. 
We also measure inter-annotator agreement (IAA) using Fleiss Kappa \cite{fleiss1971measuring}. A Fleiss Kappa score of 0.42 was obtained, which indicates a moderate agreement.

\section{Evaluation Metrics}

In this section, we describe evaluation metrics. As our approach includes a retrieval module and a summarization controller, we evaluate both components using suitable evaluation metrics. 

\subsection{Retrieval Evaluation}
\label{subsec:retrieval}

To evaluate retrieval effectiveness, i.e., the ability of the retrieval module to identify and retrieve evidence to be used to compose the summary,
we use standard information retrieval (IR) metrics, and in particular precision@$k$ (P@$k$) and recall@$k$ (R@$k$) where $k$ denotes the cut-off.
For P@$k$ and R@$k$, we map the human-annotated judgment level 3 to positive and judgment levels 0-2 to negative and report the corresponding results for $k \in \{10, 50\}$.

\subsection{Summarization Evaluation}
\label{subsec:summeval}

\subsubsection{Lexical Metrics.}

In our experiments, we measure the accuracy of generated summaries against gold summaries.
Following previous work~\cite{laskar-etal-2020-wsl,xu-lapata-2021-generating}, we employ the commonly-used standard ROUGE-1, ROUGE-2, and ROUGE-SU4~\cite{lin-2004-rouge}, which evaluate the accuracy on uni-grams, on bi-grams, and on bi-grams with a maximum skip distance of 4, respectively.
We report the $\mathrm{F_1}$ score with a maximum length of 250 words for the metrics. 
It is worth noting that there are multiple reference summaries on our dataset, we take the average of the ROUGE scores for all summaries as the final reported ROUGE score.

\subsubsection{Semantic Metrics.}

However, ROUGE-based metrics only measure lexical overlap between generated summaries and gold summaries, failing to capture their semantic similarities. Recent semantic metrics, such as BERTScore~\cite{zhang2020bertscore} and BARTScore~\cite{yuan2021bartscore}, have been shown to be effectively correlated with human judgments~\cite{huang-etal-2023-swing}. Therefore, we also report them in our experiments. Specifically, we report the F1 score of BERTscore and use recommended \texttt{deberta-xlarge-mnli}~\cite{he2021deberta} as the backbone. For BARTScore, we use the recall version of BARTScore as it shows more effective performance in summarization tasks~\cite{yuan2021bartscore}, where we use \texttt{bart-large-cnn} as the backbone. It is worth noting that the raw scores of BARTScore are negative log probabilities that are difficult to explain, we follow \cite{shaham-etal-2022-scrolls} to normalize them to positive scores with an exponential function.

\section{Experimental Setup}
\label{sec:exp}

In this section, we describe the baseline models for comparison in our experiments. We then discuss the implementation details.

\subsection{Baselines}
\label{subsec:models}

We compare our proposed method with the following two families of baseline models:

\subsubsection{\change{Weakly Supervised} QFS Models.}

To illustrate new challenges brought by our knowledge-intensive setup, we first include two weakly supervised models designed for traditional QFS tasks:
\begin{enumerate*}[label=\arabic*)]
    \item \textsc{QuerySum}~\cite{xu-lapata-2020-coarse}: an extractive QFS model leveraging distant supervision signals from trained question answering (QA) models to select salient sentences as the summary.
    \item \textsc{MargeSum}~\cite{xu-lapata-2021-generating}: an abstractive QFS model that employs generative models trained on generic summarization resources to boost sentence ranking and summary generation, achieving competitive performance in the weakly supervised setup. 
\end{enumerate*}
We discuss the adaption of two models to our knowledge-intensive setup in the implementation details.

\subsubsection{\change{Supervised} Retrieval-Augmented Models.}

We employ supervised Retrieval-Augmented Generation (RAG) models below: 
\begin{enumerate*}[label=\arabic*)]
    \item RAG-Sequence~\cite{lewis2020rag}: a generative model in which document retrieval and summary generation are learned jointly. As RAG models have two variants including RAG-Token and RAG-Sequence, we mainly report the results of RAG-Sequence as it shows better performance in knowledge-intensive tasks. \change{Specifically, the RAG-Sequence model employs DPR~\cite{karpukhin-etal-2020-dense} for retrieval and $\mathrm{BART_{large}}$~\cite{lewis-etal-2020-bart} for generation.}
    \item Fusion-in-Decoder (FiD)~\cite{izacard-grave-2021-leveraging}: an encoder-decoder architecture which has achieved state-of-art performance in knowledge-intensive tasks~\cite{asai-etal-2022-evidentiality}. In this model, encoded representations for retrieved documents are first concatenated and then fed into the decoder to generate the output. Following the origin paper, we use $\mathrm{T5_{base}}$~\cite{raffel2020exploring} as the base model. 
\end{enumerate*}

\subsubsection{\change{LLM-based RAG Models.}}

\change{In addition to supervised RAG models, we compare our approach with LLM-based RAG models~\cite{gao2023ragsurvey}. Specifically, we employ BM25 or DPR for retrieval and GPT-3.5 for generation, referring to this baseline as NaiveRAG. Following the prompting strategy from Gao et al.~\cite{gao2023ragsurvey}, we instruct GPT-3.5 to answer a given query based on the associated retrieved documents. Notably, NaiveRAG does not require fine-tuning.}

\subsection{Implementation Details}

For retrieval models, we use the library Pyserini~\cite{lin2021pyserini} to implement BM25 and DPR.\footnote{\url{https://github.com/castorini/pyserini}}
For the summarization controller, we employ GPT-3.5 using the OpenAI API service, where we use long-context version \texttt{gpt-3.5-turbo-0613}.\footnote{\url{https://platform.openai.com}}
\change{
We utilize $3$-shot demonstrations in few-shot settings and present results for both zero-shot and few-shot settings. The temperature and top-p are set to $0.1$ and $0.95$, respectively, with a maximum output length of $400$ tokens. The number of top-$k$ retrieved documents is fixed at $50$. Notably, we employed the long-context version of GPT-3.5, which supports a maximum context window of 16,385 tokens. Given that each retrieved document contains up to 100 words, when we concatenate $50$ documents into a prompt, the total length of the prompt is roughly $7,118$ tokens on average, which is comfortably below the maximum limit.
} 

For QFS models, as they have achieved competitive performance on the QFS datasets in a weak supervision manner, we do not fine-tune them on our dataset. For inference, although they include a retrieval module that performs in-domain evidence estimation, their retrieval models are not designed to handle large-scale, open-domain knowledge bases, such as \corpusext{}. Therefore, we first retrieve top-$1000$ documents using BM25 for each query and then feed them as inputs to their customized sentence extraction modules. We follow the original papers~\cite{xu-lapata-2020-coarse,xu-lapata-2021-generating} to set their hyper-parameters.

For retrieval-augmented models, as they are originally designed for short-form open-domain QA~\cite{chen-etal-2017-reading}. For fair comparisons, we fine-tune them on our training set. We follow the original papers~\cite{lewis2020rag,izacard-grave-2021-leveraging} to set their associated hyper-parameters and training setups. For inference, we set the maximum decoded length and beam size to 250 and 5, respectively. All experiments are conducted on a single A6000 GPU.

\begin{table}[t]
\centering
\caption{Retrieval evaluation of different models on the test set of our dataset over three knowledge corpora. P@$k$ and R@$k$ denote precision and recall regarding cutoff $k$. The best results are in \textbf{bold}. $^*$means that the improvement is statistically significant (a two-paired t-test with p-value < 0.01).}
\label{tab:retrieval}
\begin{tabularx}{0.53\textwidth}{l *{5}{l}}
\toprule[1pt]
Corpus & Model  & P@10 & P@50 & R@10 & R@50 \\
\hline
\multirow{2}{*}{\corpusint{}} & BM25 & \textbf{16.0}$^*$  & \textbf{12.0} & \textbf{8.2}$^*$ & \textbf{30.2}$^*$ \\
& DPR & 11.6 & 11.5 & 6.0 & 27.8 \\
\hline
\multirow{2}{*}{\corpusext{}} & BM25 & 12.7  & 8.6 & \textbf{7.0}  & 22.2 \\
& DPR & \textbf{13.8}$^*$ & \textbf{12.6}$^*$ & 6.9  & \textbf{29.7}$^*$ \\
\hline
\multirow{2}{*}{\corpusaug{}} & BM25 & 12.2 & 8.9 & 6.8 & 23.2 \\
& DPR & \textbf{14.4}$^*$ & \textbf{12.5}$^*$ & \textbf{7.4}$^*$ & \textbf{30.3}$^*$ \\
\bottomrule[1pt]
\end{tabularx}
\end{table}

\section{Results and Analysis}
\label{sec:results}

Our experimental results primarily address the following research questions (RQs): 
\begin{description}
    \item[RQ1] What is the retrieval effectiveness of different retrieval models in the knowledge-intensive setup?
    \item[RQ2] How does our approach compare to baseline models in the knowledge-intensive setup?
    \item[RQ3] What is the effect of our knowledge-intensive setup compared to the traditional QFS?
\end{description}

\subsection{Retrieval Results}
\label{subsec:retrieval_results}

To answer \textbf{RQ1}, we evaluate the retrieval effectiveness of BM25 and DPR on the test set of our dataset among three different knowledge corpora. 
The results are shown in Table~\ref{tab:retrieval}. We find that BM25 outperforms DPR on \corpusint{}. One plausible explanation is that the internal knowledge corpus (\corpusint{}) comprises documents sourced from human-curated DUC datasets, potentially containing more relevant keywords. These keywords can be easily retrieved by BM25 which is a term-based IR method.
Conversely, DPR exhibits better performance on \corpusext{}. This is because it is pre-trained on  Wikipedia data dumps, allowing it to better capture semantic representations of Wikipedia documents, thus enhancing retrieval performance.
Moreover, we observe DPR's superior performance over BM25 on \corpusaug{}. This could be due to the high similarity between \corpusaug{} and \corpusext{}, as \corpusaug{} only contains a small fraction of documents from DUC datasets and both predominantly consist of Wikipedia documents. 
However, the fairly low precision and recall scores across all three collections underscore challenges in such large-scale knowledge retrieval. Addressing these challenges requires more research efforts to enhance retrieval performance.

\begin{table*}[t]
    \small
    \centering
    \caption{Summarization evaluation of our approach and baseline models over the three knowledge corpora. R1, R2, RS, BE, and BA  stand for ROUGE-1, ROUGE-2, ROUGE-SU4, BERTScore, and BARTScore, respectively. The best results are in \textbf{bold}. $^*$indicates that the improvement to the best baseline model is statistically significant (a two-paired t-test with p-value < 0.01).
    }
    \label{tab:main_duc}
    \begin{adjustbox}{max width=\textwidth}
    {
    \begin{tabular}{l *{5}{l} *{5}{l} *{5}{l}}
        \toprule[1pt]
        & \multicolumn{5}{c}{\corpusint{}} & 
        \multicolumn{5}{c}{\corpusext{}} &
        \multicolumn{5}{c}{\corpusaug{}} \\
        \cmidrule(lr){2-6} 
        \cmidrule(lr){7-11}
        \cmidrule(lr){12-16}
        \textbf{Model}   & R1 & R2 & RS & BE & BA & R1 & R2 & RS & BE & BA & R1 & R2 & RS & BE & BA  \\ 
        \midrule
        \textit{Weakly Supervised} \\
       \querysum{}~\cite{xu-lapata-2020-coarse}  & 36.1 & 7.5 & 12.7 & 8.5 & 32.3 & 31.1 & 4.5 & 10.2 & 2.0 & 28.9 & 32.6 & 5.5 & 11.1 & 3.0 & 29.8  \\
        \margesum{}~\cite{xu-lapata-2021-generating} & 38.0 & 9.1 & 14.3 & 11.5 & 32.8 & 34.4 & 6.5 & 12.2 & 5.9 & 30.3 & 36.7 & 8.1 & 13.5 & 8.7 & 32.1 \\
         \cdashline{1-16}[1pt/1.5pt]
        \textit{Supervised} \\
        \change{RAG-Sequence}~\cite{lewis2020rag} & 28.9 & 5.7 & 10.1 & 12.6 & 4.1 & 32.3 & 5.2 & 10.8 & 8.0 & 3.9 & 27.1 & 4.6 & 9.0 & 8.3  & 3.9  \\
        \change{FiD~\cite{izacard-grave-2021-leveraging}} \\
        \quad  \change{- BM25} & 42.4 & 11.3 & 16.5 & 21.4 & 38.1 & 38.8 & 8.4 & 14.2 & 15.7 & 35.0 & 41.4 & 10.8 & 16.1 & 20.0  & 36.8 \\
        \quad  \change{- DPR} & 41.5 & 10.7 & 15.9 & 21.4 & 39.0 & 38.6 & 8.0 & 14.1 & 15.5 & 34.1 & 40.0 & 9.4 & 15.1 & 18.0 & 36.7 \\
        \cdashline{1-16}[1pt/1.5pt]
        \change{\textit{Zero-Shot Prompted}} \\
        \change{NaiveRAG~\cite{gao2023ragsurvey}} \\
        \quad  \change{- BM25} & \change{36.4} & \change{10.5} & \change{14.4} & \change{26.8} & \change{39.5} & \change{31.3} & \change{7.0} & \change{11.4} & \change{19.6} & \change{34.9} & \change{32.8} & \change{8.1} & \change{12.4} & \change{23.4} & \change{37.2} \\
         \quad  \change{- DPR} & \change{37.1} & \change{10.4} & \change{14.7} & \change{27.3} & \change{39.5} & \change{31.7} & \change{7.1} & \change{11.4} & \change{18.3} & \change{34.8} & \change{33.7} & \change{8.4} & \change{12.6} & \change{21.8} & \change{37.1} \\
        \change{Ours} \\
        \quad \change{- BM25} & \change{43.1} & \change{11.0} & \change{16.5} & \change{25.9} & \change{40.7} & \change{37.9} & \change{7.3} & \change{13.1} & \change{19.9} & \change{34.5} & \change{39.4} & \change{8.8} & \change{14.3} & \change{22.1} & \change{37.2} \\
        \quad \change{- DPR} & \change{42.8} & \change{11.0} & \change{16.3} & \change{25.5} & \change{40.6} & \change{36.7} & \change{7.0} & \change{12.5} & \change{17.7} & \change{33.8} & \change{39.2} & \change{8.6} & \change{14.0} & \change{21.5} & \change{37.2} \\
         \cdashline{1-16}[1pt/1.5pt]
        \change{\textit{Few-Shot Prompted}} \\
        \change{NaiveRAG~\cite{gao2023ragsurvey}} \\
        \quad  \change{- BM25} & \change{41.7} & \change{11.9} & \change{16.5} & \change{24.4} & \change{41.7} & \change{35.2} & \change{7.8} & \change{12.8} & \change{17.1} & \change{35.5} & \change{37.5} & \change{9.1} & \change{14.2} & \change{20.2} & \change{37.7} \\
         \quad  \change{- DPR} & \change{42.3} & \change{11.7} & \change{16.5} & \change{25.1} & \change{41.6} & \change{34.3} & \change{7.5} & \change{12.3} & \change{16.7} & \change{35.1} & \change{36.9} & \change{9.1} & \change{13.9} & \change{19.1} & \change{37.9} \\
        \change{Ours} \\
        \quad \change{- BM25} & \change{\textbf{45.8}$^*$} & \change{\textbf{13.4}} & \change{\textbf{18.6}} & \change{\textbf{29.2}$^*$} & \change{44.7} & \change{\textbf{41.7}$^*$} & \change{\textbf{9.6}} & \change{\textbf{15.6}$^*$} & \change{22.5} & \change{\textbf{37.3}} & \change{\textbf{43.6}} & \change{\textbf{11.3}} & \change{\textbf{16.7}} & \change{\textbf{26.1}$^*$} & \change{40.8} \\
        \quad \change{- DPR} & \change{45.1} & \change{12.9} & \change{18.3} & \change{28.6} & \change{\textbf{44.8}$^*$} & \change{41.1} & \change{9.4} & \change{15.2} & \change{\textbf{22.7}$^*$} & \change{36.8} & \change{42.9} & \change{10.7} & \change{16.2} & \change{24.7} & \change{\textbf{41.7}$^*$} \\
        \bottomrule[1pt]
    \end{tabular}
    }
    \end{adjustbox}
\end{table*}

\subsection{Summarization Results}
\label{subsec:summarization_results}

To answer \textbf{RQ2}, we make a comparison between our approach and baseline models, including weakly-supervised QFS, supervised RAG models, and LLM-based RAG models. The results are shown in Table \ref{tab:main_duc}. Interestingly, we find our few-shot prompted approach variant surpasses all baseline models across all evaluation metrics, especially in terms of semantic metrics.
\change{
For instance, our best approach variant achieved a BERTScore of $29.2$, surpassing the best-performing NaiveRAG, which had a BERTScore of $25.1$. This result underscores the superiority of our proposed approach.
Additionally, our approach consistently outperforms baselines across three different knowledge corpora and demonstrates robustness across various retrieval models.
Furthermore, when comparing zero-shot and few-shot settings, we find that the performance of our approach in the few-shot setting significantly exceeds that in the zero-shot setting. This indicates that few-shot demonstrations have a highly positive effect on model performance.
}
Lastly, our results reveal that fine-tuned FiD models outperform QFS models, indicating the advantages of fine-tuning on our training data. However, we also observe that the performance of the supervised RAG-Sequence model was comparatively lower, potentially due to the limited size of the training data.

\begin{table*}[t]
\centering
\caption{Performance comparison of our approach variant with BM25 over the original document sets (\origin{}) and our three knowledge corpora.}
\label{tab:qfs_duc}
\begin{tabularx}{0.85\textwidth}{l *{5}{c}}
\toprule[1pt]
Corpus  & ROUGE-1 & ROUGE-2 & ROUGE-SU4 & BERTScore & BARTScore \\
\hline
\origin{} & 49.8 & 17.3 & 21.9 & 32.5 & 47.6  \\
\corpusint{} & 45.8 & 13.4 & 18.6 & 29.2 & 44.7    \\
\corpusext{} & 41.7 & 9.6 & 15.6 & 22.5 & 37.3  \\
\corpusaug{} & 43.2 & 11.3 & 16.7 & 26.1 & 40.8  \\
\bottomrule[1pt]
\end{tabularx}
\end{table*}

\subsection{Effect of Knowledge-Intensive Setup}
\label{subsec:effect}

To answer \textbf{RQ3},
we compare the performance of our best approach variant, which utilizes BM25, across original document sets (\origin{}) and our three knowledge corpora. It is worth noting that the primary difference lies in the fact that \origin{} contains only a limited set of manually curated relevant documents (around 40 documents) from original DUC datasets, whereas our knowledge corpora, such as \corpusext{} and \corpusaug{}, consists of millions of candidate documents.
The results are shown in Table \ref{tab:qfs_duc}. We observe a significant decline in model performance when applied in the knowledge-intensive setup. For instance, when comparing \origin{} and \corpusaug{}, ROUGE-1 scores decrease by about $6$ points, dropping from $49.8$ to $43.2$, and BERTScore scores decrease by about $6$  points, dropping from $32.5$ to $26.1$.
One possible explanation is that although both knowledge corpora offer substantial evidence for summarization, \corpusaug{} contains a considerable number of irrelevant documents. Consequently, retrieval models struggle to sift through this larger pool to identify the relatively small amount of relevant documents.
This discrepancy underscores the significantly greater challenge posed by the knowledge-intensive setups compared to the traditional QFS, indicating a need for further research efforts to adapt retrieval models to these realistic scenarios.

\section{Conclusion}
\label{conclusion}

This paper introduces a novel knowledge-intensive approach to QFS, aiming to address the limitations of traditional QFS setup that relies on the availability of relevant documents. This approach is tailored for practical scenarios involving highly specialized topics, eliminating the dependence on pre-existing document sets. 
The approach efficiently retrieves potentially relevant documents from a large-scale knowledge corpus based on the query. The summarization controller combines an LLM-based summarizer with a carefully tailored prompt, ensuring the quality of the generated summary. 
To assess the effectiveness of our proposed approach compared to strong baseline models, we introduce a specialized dataset, along with human-annotated relevance labels, to facilitate comprehensive retrieval and summarization evaluation.
Extensive experiments demonstrate the superior performance of our method, particularly its ability to generate accurate summaries without relying on the availability of relevant documents initially. This underscores the versatility and practical applicability of our approach across diverse query scenarios, thereby contributing to advancements in QFS research.

\bibliographystyle{splncs04}
\bibliography{custom}

\end{document}